\documentclass[sigconf]{acmart}
\AtBeginDocument{%
  }

\copyrightyear{2026}
\acmYear{2026}
\setcopyright{usgov}
\acmConference[SIGSPATIAL '26]{The 34th ACM International Conference on Advances in Geographic Information Systems}{November 03--06, 2026}{Riverside, CA, USA}
\acmBooktitle{The 34th ACM International Conference on Advances in Geographic Information Systems (SIGSPATIAL '26), November 03--06, 2026, Riverside, CA, USA}
\acmDOI{10.1145/3841645.3843380}
\acmISBN{979-8-4007-2950-8/2026/11}

\usepackage{tcolorbox}
\newcommand{\highlightblue}[1]{\textcolor{blue}{\textbf{#1}}}
\newcommand{\highlightmagenta}[1]{\textcolor{magenta}{\textbf{#1}}}
\newcommand{\highlightorange}[1]{\textcolor{orange}{\textbf{#1}}}
\newcommand{\highlightteal}[1]{\textcolor{teal}{\textbf{#1}}}
\newcommand{\highlightcyan}[1]{\textcolor{cyan}{\textbf{#1}}}
\usepackage{multirow}
\usepackage{orcidlink}

\begin{document}

\title{Electronic Navigational Chart Change Classification}

\author{Jacob Arndt} 
\email{arndtjw@ornl.gov}
\affiliation{%
  \institution{Oak Ridge National Laboratory}
  \city{Oak Ridge}
  \state{Tennessee}
  \country{USA}
}

\author{Abhishek Potnis} 
\email{potnisav@ornl.gov}
\affiliation{%
  \institution{Oak Ridge National Laboratory}
  \city{Oak Ridge}
  \state{Tennessee}
  \country{USA}
}

\author{Alexandre Sorokine} 
\email{sorokina@ornl.gov}
\affiliation{%
  \institution{Oak Ridge National Laboratory}
  \city{Oak Ridge}
  \state{Tennessee}
  \country{USA}
}


\begin{abstract}

Electronic Navigational Charts (ENCs) are geospatial vector datasets used in maritime navigation systems that represent hydrographic and navigational information such as depths, navigational aids, traffic schemes, and hazards. A major challenge for hydrographic offices is determining whether a given chart change poses a critical or non-critical risk to maritime safety. Existing workflows rely heavily on manual review and verification, which is labor-intensive, scales poorly with the volume of incoming chart updates, and introduces inter-analyst inconsistencies. To address this challenge, we propose a method for automated classification of ENC changes. We establish a baseline encoding scheme to translate complex vector data changes into a structured tabular format for classification models. The two crucial components of the encoding scheme include a spatial context encoder to enrich the change representations with surrounding geographic features, and an ENC attribute encoder to represent nuanced attribute-value descriptions of the modified objects. We evaluate the proposed approach across two distinct operational datasets, comprising 1,308 chart pairs containing over 100,000 individual chart modifications. Tuned gradient-boosted trees leveraging the proposed encoding schemes achieve accuracies of 90\% and 94\% on the two datasets, yielding a 5-7\% improvement over default hyperparameterized models trained on encodings without spatial context and attribute embeddings. These results demonstrate the viability of integrating machine learning into operational geospatial pipelines to improve ENC maintenance and enhance maritime safety. Finally, our experiments demonstrate the effectiveness of simple location and spatial aggregation methods, providing a foundation for evaluating more sophisticated spatial representation learning techniques for this application.

\end{abstract}




\maketitle

\section{Introduction}

\begin{figure}[t!]
    \centering
    \includegraphics[width=0.98\linewidth]{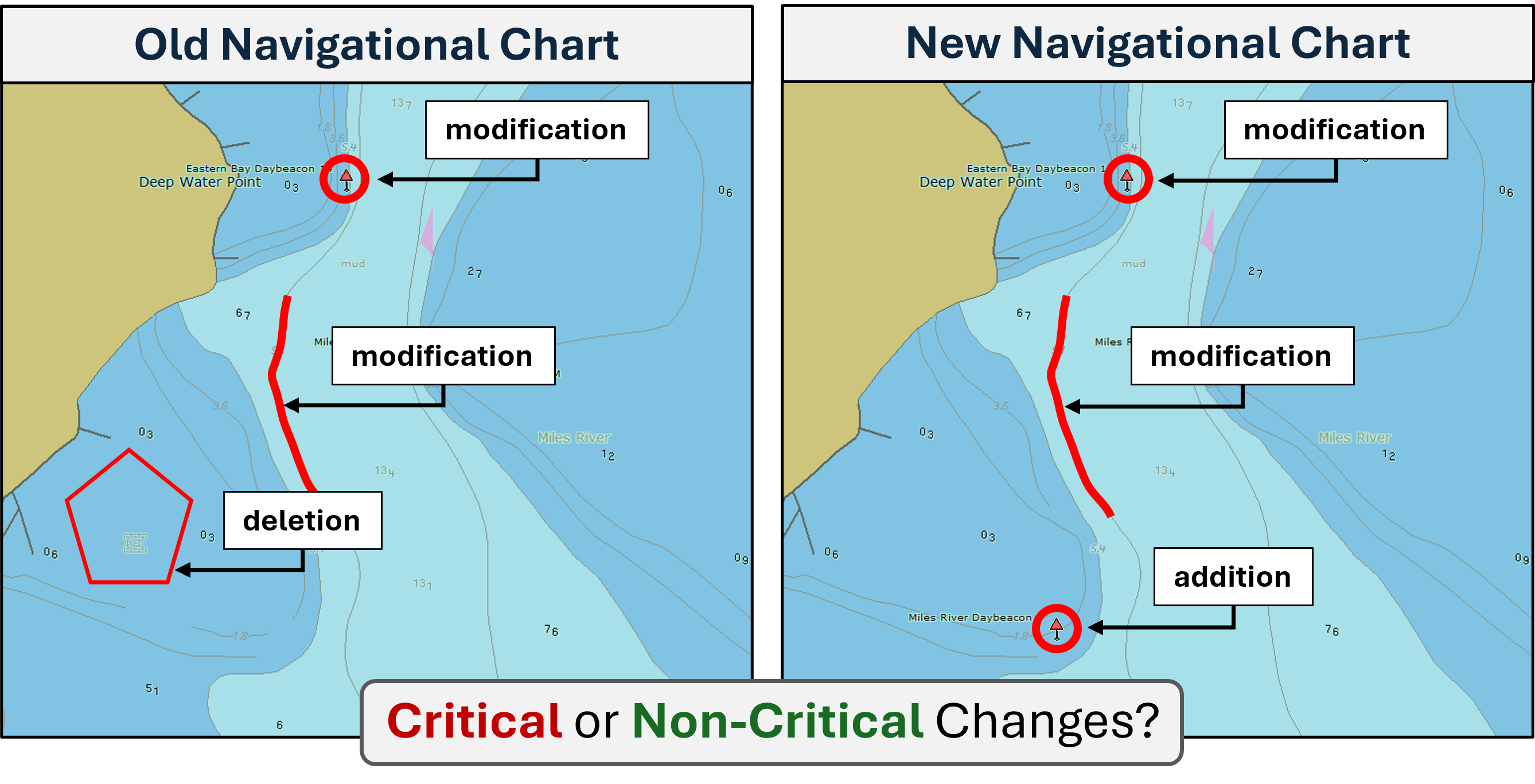}
    \caption{Changes in ENCs must be classified in terms of their significance to navigational safety. ENC changes can include additions or deletions of ENC objects, or modifications to an ENC object's attributes, geometry, relationships, or spatial context.}
    \label{fig:description}
\end{figure}

Maintaining Electronic Navigational Charts (ENCs) is a requirement and significant challenge for many hydrographic offices \citep{Masetti2018}. New incoming data in the form of hydrographic surveys or updated ENCs are received at an increasingly rapid rate and must be assessed for their potential impact to navigational safety at sea. Requirements and guidance for maintenance is described in detail in the International Hydrographic Organization's (IHOs) Standard Reference S-4. 
A key step in the ENC maintenance workflow is determining whether observed differences between new data and existing data represent critical or non-critical changes (Figure~\ref{fig:description}). 
Critical changes, those that are urgent and pose an immediate risk to navigational safety, must be promptly incorporated into the ENC and disseminated using appropriate mechanisms while non-critical changes must be recorded and queued for the next appropriate chart revision. Ensuring that this process is efficient and accurate is absolutely mandatory, as delays or misclassifications can result in serious consequences and risk to safety. 

Classifying ENC change criticality is currently performed manually by domain experts with extensive knowledge of maritime data, cartography, and maritime navigation. Upon receiving an ENC change, a typical human review must account for multiple factors, including: 1) the type of change (e.g., addition, deletion, modification), 2) the affected ENC object class (e.g., light, sounding, depth contour), 3) object attributes (e.g., sector limits, depth value, category of light), 4) spatial relationships (e.g., proximity to other objects, shared boundaries), 5) geometry modifications, 6) collection membership (associations, aggregations, master-slave), and 7) chart metadata (e.g., navigational purpose, compilation scale). To further illustrate the diversity of information and complexity of the decision-making, consider the example in Figure~\ref{fig:iho-change-example}. 

\begin{figure}[h!]
  \centering
  \footnotesize
  \begin{minipage}{0.98\linewidth}
    \begin{tcolorbox}[colback=gray!5!white, colframe=gray!75!black]
      \highlightblue{Addition} or \highlightblue{modification} of a \highlightorange{light} object's \highlightmagenta{sector} attributes. Criticality of the change is determined by the \highlightteal{spatial proximity} of the \highlightmagenta{sector limit} to a \highlightorange{danger} object. Movement of the sector limit must be plottable by the chart user, which depends on the \highlightcyan{chart scale} and the \highlightmagenta{range} of the light. This is unlikely to be less than 1 degree for long-range lights and less than 3 degrees for short-range lights. -- \textit{S-4 Part B, Section-600 Chart Maintenance} \\ 

      Types of information: \highlightblue{Change type}, \highlightorange{ENC object class}, \highlightmagenta{ENC attributes}, \highlightteal{Spatial relationships}, \highlightcyan{Chart metadata}
    \end{tcolorbox}
  \end{minipage}
  \caption{Summarized excerpt providing guidance on change criticality for a light sector. Color-coding indicates types of information considered in determining change criticality.}
  \label{fig:iho-change-example}
\end{figure}


      

One approach to automate ENC change classification is to program expert-defined rules for each possible change scenario. This can be used for simple change scenarios, however, the method does not generalize to the full complexity of ENC data and its maintenance. ENCs contain hundreds of object classes, attributes, and attribute values of diverse types along with multiple geometry types, scale dependencies, defined object-to-object semantic relationships, spatial relationships, and topological rules. Additionally, hydrographic offices often maintain different ENC products that require different criticality rules which further complicates rule development, management, and application. As a result of this complexity, many changes still require human review. 

To address these challenges, we introduce a machine learning framework for the automated classification of ENC changes as critical or non-critical. We propose encoding methods that translate complex vector chart modifications into structured tabular representation suitable for machine learning models. Our approach features a spatial context encoder to capture surrounding geographic objects and an attribute encoder to ingest nuanced object properties. We evaluate our encoding methods along with several baseline classification models across two operational datasets and demonstrate the improvement provided by these spatial and attribute representations. To the best of our knowledge, this is the first work to apply machine learning to automated ENC change classification.

The main contributions of this paper are:
\begin{itemize}
    \item We introduce ENC change classification as a spatial data science and machine learning problem.
    \item We propose a set of ENC change data encodings that include both spatial context and attribute representations.
    \item We develop evaluation protocol to benchmark data encodings and classification models for classifying ENC changes as critical or non-critical.
\end{itemize}

\section{Background}
The S-57 data standard \citep{IHO_S57}
defines a highly structured theoretical data model and data structure to permit the exchange of digital hydrographic data. Within the S-57 data standard, the ENC product specification ensures consistency in the production of ENC data and its usage in on-vessel navigation systems. ENCs are complex and rich spatial datasets organized into ``cells'' consisting of: 1) cartographic scale dependency, 2) diverse geometry types (points, lines, areas), 3) hierarchical semantic relationships (e.g. master-slave relationships between objects), 4) spatial topological relationships (e.g. depth contour touches depth area), 5) diverse attribute types (e.g. free-text, number, categorical, list), 6) specific object-attribute relationships (e.g. category of light attribute is applicable to light objects, but not applicable to coastline objects) and 7) consistent updates (ENCs are constantly updated to reflect real-world conditions).

\section{Methods}


\subsection{Preliminaries}
Given as input an ENC change instance \( \mathcal{I} = (C_{\text{old}}, C_{\text{new}}, \Delta) \), our objective is to predict a \textit{critical} or \textit{non-critical} class label. An ENC change instance consists of an existing ENC cell \( C_{\text{old}} \), a new ENC cell \( C_{\text{new}} \), and an identified change \( \Delta = (\Delta_{\text{old}}, \Delta_{\text{new}}) \). Each ENC cell includes ENC objects \( \mathcal{O} = \{g_1, g_2, \dots, g_n\} \), where each object \( g_i \) is described by an ENC object class \( g_i^{\text{class}} \) (e.g. sounding, lighthouse, coastline, bouy), a 2D geometry \( g_i^{\text{geom}} \) such as a point, line, or polygon, a set of key–value(s) attributes \( g_i^{\text{attr}} = \{(a_1, v_1), \dots, (a_k, v_k)\} \) describing the object, and S-57 relationships (\( g_i^{\text{rel}} \subseteq \mathcal{R} \)) (e.g. aggregations, associations, master-slave) linking the object to other ENC objects in the cell.

The identified change  \( \Delta = (\Delta_{\text{old}}, \Delta_{\text{new}}) \) contains information about changed ENC objects in the old and new ENC cells. \( \Delta_{\text{old}} = \{ o_1^{\text{old}}, o_2^{\text{old}}, \dots \} \subseteq C_{\text{old}} \) contains references to ENC objects that have been either deleted or modified in the old ENC cell, and \( \Delta_{\text{new}} = \{ o_1^{\text{new}}, o_2^{\text{new}}, \dots \} \subseteq C_{\text{new}} \) contains references to ENC objects that have been either added or modified in the new ENC cell. All ENC objects referenced in an ENC change \(\Delta\) share the same ENC object class. Five change types are possible depending on the number of objects in the identified change. These change types include \textit{addition}, \textit{deletion}, \textit{modification}, \textit{multi-feature modification}, and \textit{duplicate feature modification}. For example, a \textit{deletion} change type will consist of one object in \( \Delta_{\text{old}} \) and zero objects in \( \Delta_{\text{new}} \).

\subsection{ENC Change Encoding}

We encode ENC changes into numeric feature vectors \(h_{change}\) designed to be representative of the identified change. Our encoding options include ENC object class, the type of change, ENC relationships for the change objects, the ENC attributes of the change objects, and the spatial context of the change object in the old and new ENC cells.

\textbf{Object class.} We use a simple one-hot encoding to represent the ENC object class for the change. With 172 possible ENC object categories, the ENC object class one-hot encoder \( E_{\text{class}} \) produces a binary vector \( h_{\text{class}} \) with dimension 172.


\textbf{Change type.} We use a one-hot encoding to represent ENC change type. The ENC change type one-hot encoder \( E_{\text{type}} \) produces a binary vector \( h_{\text{type}} \) with dimension 5.



\textbf{Relationships.} To obtain a representation of S-57 ENC standard relationships for an ENC change, we develop a simple aggregation function. We consider five relationship types including \textit{aggregation peers}, \textit{association peers}, \textit{slave peers}, \textit{slave objects}, and \textit{master object}. We define \(\mathcal{R}_r(x)\) as the set of ENC objects related to the change object \( x \) via relationship \(r\). For each relationship type \(r\), we one-hot encode the object class of all objects in \(\mathcal{R}_r(x)\) and aggregate the resulting vectors using an element-wise sum. To form the final ENC relationships vector \( h_{\text{rel}} \) for the ENC change, we concatenate the vectors obtained from each relationship type \(r\) and for the old and new contexts resulting in \( h_{\text{rel}} \) of dimension 1720.

\begin{align}
h_r(x) = \sum_{x_{i} \in \mathcal{R}_r(x)} E_{\text{class}}(x_{i}),
\end{align}


\textbf{Spatial context.} We develop a simple spatial context encoder that encodes and aggregates ENC objects that neighbor a given ENC change instance. Specifically, we treat the identified change object as a target node and its neighboring ENC objects as source nodes in a spatial neighborhood. We define $\mathcal{N}(\Delta_\text{c})$ as the set of ENC objects within a spatial neighborhood of the change for context $c \in \{\text{old}, \text{new}\}$. For an ENC change instance $\Delta$, the old neighborhood $\mathcal{N}(\Delta_\text{old})$ and new neighborhood $\mathcal{N}(\Delta_\text{new})$ are defined as all ENC objects intersecting a 500-meter radius buffer around the centroid of the change instance geometry. Objects in $\mathcal{N}(\Delta_\text{old})$ are queried from the old ENC cell, while objects in $\mathcal{N}(\Delta_\text{new})$ are queried from the new ENC cell.

For each neighbor $x_i \in \mathcal{N}(\Delta_c)$, the message function is defined as a static one-hot encoding of the ENC object class, $\mathrm{E_{class}}(x_i)$. The target node then computes its context representation $h_c$ by applying an element-wise sum as the aggregation function, accumulating the incoming messages:

\begin{equation}
h_{c} = \sum_{x_i \in \mathcal{N}(\Delta_{c})} \mathrm{E_{class}}(x_{i})
\end{equation}

This aggregation yields a localized bag-of-words (or histogram) representation of ENC object classes surrounding the change, a simple case of an aggregation location encoder \cite{mai2022review} inspired by the graph neural network framework. Finally, the spatial representations from both old and new contexts are concatenated to form the complete spatial context encoding $h_{\text{context}}$ of dimension 344.





\textbf{Attributes.} For attribute encoding we convert all key-value(s) attributes for an ENC object into strings and concatenate them. All numeric codes and values are translated into their natural language description found in Appendix A of the S-57 data standard. We separate each key and value with a colon and space and separate each key-value pair with a comma and space. For example, an attribute set $\{\text{BOYSHP}: 4, \text{COLOUR}: \text{[}3\text{]}, \text{CATLAM}: 2, \text{OBJNAM}: \text{``Little peconic bay lighted buoy 18''}\}$ is translated into: ``Buoy shape: Pillar, Colour: Red, Category of lateral mark: Starboard-hand lateral mark, Object name: Little peconic bay lighted buoy 18''. After these strings are constructed, we tokenize and embed them using the DistilBERT \citep{sanh2019distilbert}
transformer architecture pretrained on Wikipedia and the Toronto Book Corpus. 
We embed the attributes in this way for the the old change object and new change object and concatenate the resulting vectors together to form a single attribute vector $h_{attr}$ with dimension 1536.


Finally, our complete ENC change encoding is a concatenation of $h_{\text{class}}$, $h_{\text{type}}$, $h_{\text{rel}}$, $h_{\text{context}}$, and $h_{\text{attr}}$ resulting in the final change vector $h_{change}$ of dimension 3777.


\subsection{Classification Models}
We use XGBoost (XGB) 
as our primary baseline model to classify each identified change as critical or non-critical. XGBoost, and more broadly, ensemble decision tree based models achieve strong performance in tabular machine learning on numerous datasets \citep{grinsztajn2022tree} and are often chosen as a competitive baseline for benchmarking \citep{gorishniy2021revisiting}. In addition to XGBoost, we train and evaluate several other deep learning and non-deep learning models including logistic regression (LR), 
random forest (RF), 
a multilayer perceptron (MLP), 
and a ResNet. 

\subsection{Evaluation}

\textbf{Datasets.} We consider two datasets in our evaluation. The first dataset, hereafter referred to as the \textit{Critical/Non-Critical} dataset, consists of ENC changes that have been automatically categorized as critical or non-critical by a programmed rule-based system. The second dataset, hereafter referred to as the \textit{Eyes-On} dataset, consists of ENC changes that do not have a programmed rule. Each change in this dataset was reviewed by human analysts and categorized as \textit{critical} or \textit{non-critical} depending on the change's significance to navigation. These ENC changes are more complex than those found in the \textit{Critical/Non-Critical} dataset, requiring multi-step decisions, and including scenarios where written rules do not exist and are not easy to formulate. There are approximately 140,000 change instances in the \textit{Critical/Non-Critical} dataset and approximately 9,000 change instances in the \textit{Eyes-On} dataset. The ENC change instances in both datasets are derived from 356 ENC cells and a total of 1308 old-to-new chart pairs covering waters in the United States, Canada, Australia, New Zealand, Norway, Netherlands, and Germany.

\textbf{Metrics and Protocol.} All models are evaluated using standard metrics for binary classification including overall accuracy and macro F1-score. We use five-fold cross-validation, partitioning each dataset into five distinct train-validation splits and reporting the average performance across all folds. We train, evaluate, and report results using both the \textit{Critical/Non-Critical} and \textit{Eyes-On} datasets.

\section{Experiments and Results}
\textbf{Spatial Context and Attribute Encoding.} Here, we compare results obtained using a baseline data encoding consisting of only ENC object class, ENC relationships, and ENC change type to results obtained with the improved ENC change encoding that includes attributes and spatial context (Table~\ref{tab:baseline-encoding}).


\begin{table}[h!]
    \caption{XGBoost trained on baseline ENC change encoding and performance impact with spatial context and attribute encoding.}
    \label{tab:baseline-encoding}
    \centering
    \small
    \resizebox{\linewidth}{!}{%
        \setlength{\tabcolsep}{3pt}
        \begin{tabular}{lcccc}
        \toprule
        & \multicolumn{2}{c}{Eyes-On}
        & \multicolumn{2}{c}{Crit./Non-Crit.} \\
        \cmidrule(lr){2-3} \cmidrule(lr){4-5}
        Encoding & Accuracy & F1-score & Accuracy & F1-score \\
        \midrule
        Baseline & .828 $\pm$ .004 & .790 $\pm$ .005 & .891 $\pm$ .002 & .891 $\pm$ .002 \\
        \hline
        + Attributes & .857 $\pm$ .007 & .837 $\pm$ .008 & .922 $\pm$ .001 & .922 $\pm$ .002 \\
        + Spatial Context & .901 $\pm$ .004 & .889 $\pm$ .004 & .900 $\pm$ .002 & .900 $\pm$ .002 \\
        + Spatial Context and Attributes & .892 $\pm$ .005 & .879 $\pm$ .005 & .935 $\pm$ .001 & .935 $\pm$ .001 \\
        \bottomrule
        \end{tabular}
    }
\end{table}

\textbf{Hyperparameter Tuning.} In Table~\ref{tab:hyperparamter-tuning}, we present results comparing XGBoost trained with default hyperparameters to XGBoost after hyperparameter tuning. For tuning, we use a tree-structured parzen estimator and median pruning, implemented using the Optuna library \cite{akiba2019optuna}, and use a similar search space as \cite{grinsztajn2022tree, gorishniy2021revisiting} for the models and training. For both default and tuning, we use the complete ENC change encoding that includes spatial context and attribute encoding. As shown in Table~\ref{tab:hyperparamter-tuning}, careful hyperparameter tuning of the model improves classification accuracy and F1-score on both datasets. The effect of hyperparameter tuning is consistent with results in \cite{grinsztajn2022tree} where they advocate for carefully tuned baselines.


\begin{table}[h!]
    \caption{Hyperparameter tuning results with XGBoost. Results are reported as \textit{Overall Accuracy}/\textit{F1-Score}.}
    \label{tab:hyperparamter-tuning}
    \centering
    \small
    \setlength{\tabcolsep}{3pt}
        \begin{tabular}{lcccc}
        \toprule
        & \multicolumn{2}{c}{Eyes-On}
        & \multicolumn{2}{c}{Crit./Non-Crit.} \\
        \cmidrule(lr){2-3} \cmidrule(lr){4-5}
        Model & Default & Tuned & Default & Tuned \\
        \midrule
        XGB & .892/.879 & .903/.891 & .935/.935 & .942/.942 \\
        \bottomrule
        \end{tabular}
\end{table}

\textbf{Comparing Baseline Models.} Here, we compare XGBoost with the other baseline models for ENC change classification. We include two simple rule-based majority vote classifiers as naive baselines, establishing a lower bound for expected performance. These include \textit{Critical Always}, which predicts the critical class for every change instance, and \textit{Majority by ENC Class}, which predicts the critical class if that ENC object class has more critical examples than non-critical in the training dataset, and the non-critical class otherwise. For the \textit{Critical/Non-Critical} dataset, we also include the performance of the \textit{Rules} system, the deterministic procedure used to auto-classify changes. These rules generated the ground-truth labels for the \textit{Critical/Non-Critical} dataset and therefore achieves 100\% accuracy and serves only as a point of reference rather than a meaningful predictive model.

Results are reported for default and tuned baseline models on both datasets in Table~\ref{tab:combined-baseline-results}. All baseline models when trained on the complete ENC data encoding can effectively distinguish between critical and non-critical changes with accuracy greater than 85\%. Furthermore, these models achieve much better performance than the two naive rule-based classifiers.

\begin{table}[h!]
    \caption{Evaluation of baseline models. Results are reported as \textit{Overall Accuracy}/\textit{F1-Score}.}
    \label{tab:combined-baseline-results}
    \centering
    \small
    \setlength{\tabcolsep}{3pt}
        \begin{tabular}{lcccc}
        \toprule
        & \multicolumn{2}{c}{Eyes-On}
        & \multicolumn{2}{c}{Crit./Non-Crit.} \\
        \cmidrule(lr){2-3} \cmidrule(lr){4-5}
        Model & Default & Tuned & Default & Tuned \\
        \midrule
        LR     & .857/.838 & .860/.841 & .888/.888 & .889/.889 \\
        RF     & .885/.871 & .894/.881 & .933/.932 & .939/.938 \\
        XGB    & .892/.879 & .903/.891 & .935/.935 & .942/.942 \\
        MLP    & .890/.876 & .898/.886 & .918/.918 & .920/.920 \\
        ResNet & .892/.877 & .904/.893 & .928/.928 & .929/.929 \\
        \midrule
        Critical Always$^{a}$ & .654/.395 & -- & .463/.317 & -- \\
        Majority by ENC$^{a}$ & .829/.794 & -- & .801/.798 & -- \\
        Rules$^{b}$ & -- & -- & 1.000/1.000 & -- \\
        \bottomrule
        \end{tabular}
    \vspace{2pt}\\
    \footnotesize
    $^{a}$ Naive majority vote classifiers.
    $^{b}$ Rule-based system.
\end{table}

\section{Conclusion}

Maintaining ENCs is a critical and resource-intensive task to accurately reflect navigation and safety-critical information to mariners. In this work, we introduced ENC change classification as a novel Geospatial AI task and introduced an encoding method that successfully translates complex bi-temporal ENC vector data into a structured format for machine learning. By integrating spatial context and attribute encodings into the ENC change representation and hyperparameter-tuning an XGBoost classifier, we achieved 90\% and 94\% overall accuracy in classifying ENC changes as critical or non-critical across two operational datasets. These baselines provide a strong foundation for evaluating more sophisticated spatial representation learning techniques. Future work will explore location encodings for heterogeneous geometry types, complex spatial aggregation methods, and spatial graph neural networks to further enhance maritime safety pipelines.

\begin{acks}
We acknowledge that this manuscript has been authored by UT-Battelle, LLC under Contract No. DE-AC05-00OR22725 with the U.S. Department of Energy. The United States Government retains and the publisher, by accepting the article for publication, acknowledges that the United States Government retains a non-exclusive, paid-up, irrevocable, world-wide license to publish or reproduce the published form of this manuscript, or allow others to do so, for United States Government purposes. DOE will provide public access to these results of federally sponsored research in accordance with the DOE Public Access Plan (http://energy.gov/downloads/doe-public-access-plan). Research sponsored by the Laboratory Directed Research and Development Program of Oak Ridge National Laboratory, managed by UT-Battelle, LLC, for the U. S. Department of Energy.
\end{acks}

\bibliographystyle{ACM-Reference-Format}
\bibliography{references}

\end{document}